\documentclass[letterpaper, 10 pt, conference]{ieeeconf}  % Comment this line out if you need a4paper

\IEEEoverridecommandlockouts                              % This command is only needed if 
\usepackage{graphicx} % for pdf, bitmapped graphics files
\usepackage{placeins} % If you need \FloatBarrier
\usepackage{hyperref}
\usepackage{siunitx}
\usepackage{float}
\usepackage{booktabs}
\usepackage{amsmath,amssymb,amsfonts}
\usepackage{algorithmic}
\usepackage{textcomp}
\usepackage{xcolor}
\usepackage{longtable}
\usepackage{caption}
\usepackage{gensymb}
\usepackage[square, numbers, sort&compress]{natbib}

\newcommand{\mypara}[1]{\par\vspace*{1mm}\noindent\textbf{{#1}}}

\usepackage{pifont}% http://ctan.org/pkg/pifont
\newcommand{\cmark}{\ding{51}}%

\title{The Duke Humanoid: Design and Control For Energy-Efficient Bipedal Locomotion Using Passive Dynamics}

\author{Boxi Xia, Bokuan Li, Jacob Lee, Michael Scutari, Boyuan Chen% <-this % stops a space
\thanks{*This work is supported by DARPA FoundSci HR00112490372, DARPA TIAMAT HR00112490419, ARL STRONG W911NF2320182 and W911NF2220113. All authors are from Duke University.}}

\begin{document}

\maketitle

\begin{abstract}
We present the Duke Humanoid, an open-source 10-degrees-of-freedom humanoid, as an extensible platform for locomotion research. The design mimics human physiology, with symmetrical body alignment in the frontal plane to maintain static balance with straight knees. We develop a reinforcement learning policy that can be deployed zero-shot on the hardware for velocity-tracking walking tasks. Additionally, to enhance energy efficiency in locomotion, we propose an end-to-end reinforcement learning algorithm that encourages the robot to leverage passive dynamics. Our experimental results show that our passive policy reduces the cost of transport by up to $50\%$ in simulation and $31\%$ in real-world tests. Our website is \url{http://generalroboticslab.com/DukeHumanoidv1/}.
\end{abstract}

\section{Introduction}

Recent advances in robotics have enabled humanoids to perform impressive dynamic motions, such as walking, high-speed running, and backflipping \cite{bostondynamics24atlas, agility24digit, unitree24g1,tesla24bot}. However, many commercially available humanoids are closed systems with limited customizability and access to low-level sensor and actuator interfaces. Some recent efforts open-source the software \cite{liao2024berkeley,li2023dynamic}, while others offer open-source hardware but are limited in their ability to perform dynamic motions due to the use of high-reduction-ratio servo motors \cite{lapeyre2014poppy,ficht2018nimbro,metta2010icub}.

The demand for customizable and high-performance humanoid platforms in the research community is evident. Yet, there remains a scarcity of open-source humanoid hardware capable of performing dynamic locomotion. Such platforms are essential for enabling transparent access, customization, and investigation of low-level hardware interactions, all of which are critical to advancing humanoid research.

\begin{figure}[!t]
\centering
\includegraphics[trim={ 0  0  0 0.1cm},clip,, width=0.8\columnwidth]{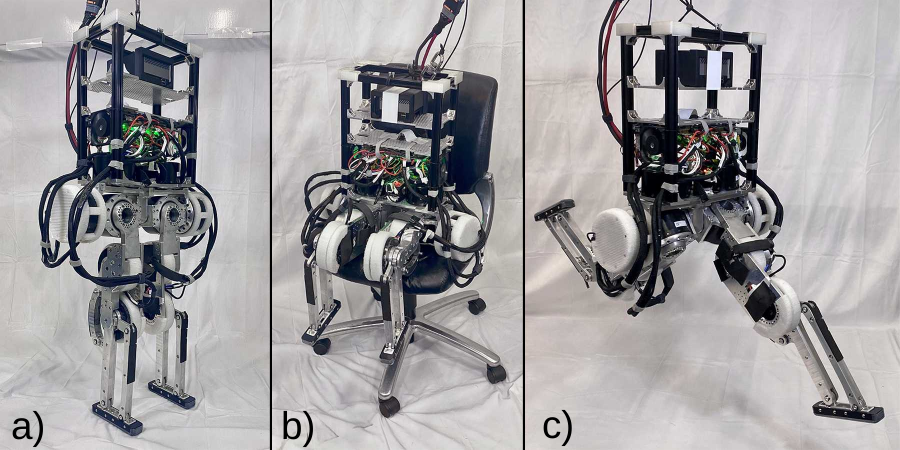}
\caption{\textbf{Duke Humanoid v1.0}: a) The frontal plane symmetry of the hip enables static standing with straight knees. b) and c) Additional poses demonstrating the robot's range of motion.}
\label{fig:titlegraphics}
\vspace*{-8mm}
\end{figure}

This paper introduces an open-source humanoid platform (Fig.~\ref{fig:titlegraphics}), designed to address the lack of open-source hardware capable of dynamic motion. The platform is customizable, allowing researchers to integrate design changes and co-optimize hardware and control systems. Our hardware design mimics human physiology by referencing the ratios of the femur, tibia, foot length, and stance width. The humanoid has a nearly symmetric weight distribution, naturally supporting static straight-knee standing. Additionally, we demonstrate the platform's capabilities by deploying a reinforcement learning (RL) walking policy, trained in simulation, to the physical system without fine-tuning. Both hardware and software are open-sourced.

With our platform, we further study a fundamental question in humanoid locomotion: How can robots achieve energy-efficient human-like walking? Many humanoids are designed with human-like morphology, which simplifies control but requires continuous active joint actuation, leading to inefficient energy use. This inefficiency results in frequent charging and battery replacement, leading to additional operational costs. Addressing this bottleneck requires both hardware innovations and improved control strategies.

Unlike most humanoids that rely on consistent active motor control in locomotion, humans leverage passive dynamics in their gait. Human walking requires minimal muscular effort for leg swinging by utilizing pendulum-like motion \cite{kuo2007six,cavagna1976sources}. Some bipedal robots use passive dynamics to achieve human-level walking efficiency while requiring little to no power input \cite{garcia2000efficiency,ohta2001passive,collins2005efficient,wisse2006design}. However, their mass distribution is specifically tuned for passive walking, and any customization, such as adding a payload, may compromise their ability to walk effectively and efficiently. Previous work \cite{garcia2000efficiency} indicates that straight-knee designs promote efficient walking without active control. These designs can maintain static equilibrium in a standing pose by fully locking the stance leg, while allowing a slight unlock of the knee in the swing leg. Our mechanical design satisfies this criterion through our near-symmetrical hip arrangement.

In parallel, research on energy-efficient legged robot control has proposed learning gait patterns \cite{yang2022fast}, using compliant reinforcement learning \cite{hartmann2024deep}, and optimizing the center-of-mass trajectory \cite{kormushev2019learning}. Although end-to-end deep reinforcement learning has achieved state-of-the-art locomotion \cite{hwangbo2019learning, miki2022learning, lee2020learning}, reinforcement learning based controller design has not yet sufficiently explored incorporating passive dynamics for bipedal locomotion within their framework. While passive dynamics has been widely studied in traditional controller design, its potential application within learning-based controller remains relatively untapped.

In this paper, we aim to enhance the energy efficiency of humanoids through an early exploration of utilizing passive dynamics in the end-to-end reinforcement learning framework. Our key idea is to explicitly modulate the joint torques to switch between passive and active control during policy learning. The simulation result shows that our passive policy can achieve up to a $50\%$ increase in energy efficiency at low-speed walking ($<0.5\mathrm{m/s}$) compared to a baseline RL policy. We validate the effectiveness of our passive action policy by deploying it on our physical platform, showing a $31\%$ increase in energy efficiency.

Our main contributions are twofold: (1) we present an open-source humanoid platform for dynamic locomotion research, and (2) we demonstrate that the use of passive dynamics in the end-to-end reinforcement learning framework leads to enhanced energy efficiency in bipedal locomotion.
%demonstrating a $31\%$ increase in energy efficiency at low-speed walking.

\section{Related Work}

\subsection{Humanoid Platforms}

The robotics industry offers several humanoid platforms of various scales \cite{agility24digit,unitree24g1,tesla24bot,bostondynamics24atlas}. However, commercial humanoids often have limited customizability and accessibility due to closed-source hardware. This lack of openness makes it difficult for researchers to modify or experiment with the robot's low-level systems, limiting opportunities for fundamental research. We address this limitation by offering an open-source platform. Academic labs have developed humanoids for research purposes \cite{liao2024berkeley,tsagarakis2013compliant,saloutos2023design,ficht2018nimbro,carpentier2021recent,englsberger2014overview,asano2016human,ahn2023development,li2023dynamic}. While these robots may not be as polished or robust as their commercial counterparts, they offer more flexibility in customizing hardware and control, enabling researchers to push the boundaries of hardware and algorithm designs.
 
Mid-sized humanoids strike a balance between functionality and cost. Compared to adult-sized robots, they are easier to design, manufacture, and handle. While miniature robots offer cost advantages, their size limits their ability to fully replicate human locomotion. Table~\ref{table:robot_compare} compares key parameters of several mid-sized industrial and academic humanoids. Among them, our platform is open-source in both hardware and software. Our design features \SI{0.5}{\meter} legs, tall enough to access standard workbenches and tables. Similar to previous efforts \cite{li2023dynamic,saloutos2023design}, we limit the ankle's degrees of freedom (DoF) to one (plantar flexion/dorsiflexion) to reduce leg mass and inertia, allowing the motor to handle higher payloads while providing sufficient mobility for dynamic walking. Additionally, our humanoid provides a maximum torque surpassing that of most mid-sized humanoids, making it suitable for dynamic locomotion with higher torque and payload demands.

\subsection{Control Strategies for Humanoid Locomotion}

\mypara{Model-Based Control} Model-based control approaches like Zero-Moment Point (ZMP) \cite{sugihara2002real} ensure stability by maintaining the center of pressure within the support polygon. Model predictive control (MPC) \cite{scianca2020mpc} optimizes future trajectories and control actions based on a dynamic model of the robot. Whole-body control (WBC) \cite{sentis2006whole} coordinates the robot's entire body motion while accounting for its dynamics and constraints. Although highly successful, model-based control requires extensive modeling of the specific robot and often relies on numerous assumptions about its dynamics, limiting transferability to other humanoids without expert knowledge.

\mypara{Learning-Based Control} Learning-based control \cite{radosavovic2024real,gu2024advancing, seo2023deep,2018-TOG-deepMimic} has gained popularity due to its ability to operate with less prior knowledge of specific robots. One approach is to use imitation learning, which directly learns from expert gait demonstrations. However, this method requires access to expensive motion capture data and is prone to errors when the policy encounters situations that differ from the expert demonstrations. Another promising strategy is to use RL to learn control policies through trial and error by maximizing reward signals. We build our control framework based on this formulation due to its flexibility and minimal requirement of prior knowledge or expert demonstrations. When coupled with the recent development of massively parallel RL simulations \cite{rl-games2021,makoviychuk2021isaac}, this allows us to achieve effective control policies with rapid convergence.

\begin{table}[t!]
% \vspace*{-2mm}
\centering
\begin{tabular}{p{1.8cm}p{0.7cm}p{0.5cm}p{0.5cm}p{0.5cm}p{0.7cm}p{0.7cm}}
\toprule
\multicolumn{2}{l}{Robot \hfill Symmetry*} & Leg length [$\si{\meter}$]  &  Leg DoF  &  Mass [$\si{\kilogram}$] & Max HFE [N$\cdot$m] & Max KFE [N$\cdot$m] \\
 % Robot  &Symm- etry* & Leg length [$\si{\meter}$]  &  Leg DoF  &  Mass [$\si{\kilogram}$] & Max HFE [N$\cdot$m] & Max KFE [N$\cdot$m] \\
\midrule
MIT \cite{saloutos2023design}&  & 0.55 & 5 & 24& 68& 136\\
Berkeley \cite{liao2024berkeley}&  & 0.4 & 6& 16& 63& 81\\
Unitree G1 \cite{unitree24g1}&  & 0.6 & 6& 35& 88& 120\\
HECTOR \cite{li2023dynamic} &  & 0.44 & 5 & 16& 33.5& 51.9\\
iCub \cite{metta2010icub} &  & 0.4& 6& 24& 40& 40\\
COMAN \cite{tsagarakis2013compliant} &  & 0.44& 6& 55& 55& 40\\
% Xbot-S \cite{gu2024advancing}&  &  - & 6& 38& P & -& -\\
\textbf{Ours} & \cmark  & 0.5& 5 & 30 & 264 & 238\\
\bottomrule
\end{tabular}
\begin{flushleft}
Symmetry*: hip arrangement symmetry across the frontal plane
% leg length is the distance from hip to ankle
\end{flushleft}
\vspace*{-3mm}
\caption{Comparison of key parameters for mid-sized academics and industrial humanoids.}
\label{table:robot_compare}
\vspace*{-15pt}
\end{table}

\subsection{Strategies for Energy-Efficient Locomotion}

\mypara{Hardware Strategies} As discussed by Garcia et al. \cite{garcia2000efficiency}, hardware designs for passive dynamic walking often benefit from straight-knee configurations to enable efficient walking without active control. A key design requirement for passive walking is maintaining static equilibrium when standing on flat ground, which can be achieved by fully locking the stance leg while allowing the swing leg's knee to slightly unlock. Various studies \cite{collins2005efficient,ikemata08passiveknee,ohta2001passive,wisse2006design} have demonstrated the effectiveness of straight-knee designs in achieving fully or mostly passive walking. Additionally, incorporating energy exchange components can enhance energy transmission. For instance, the DURUS robot \cite{reher2016realizing} employs compression springs in the ankles, while the COMAN robot \cite{tsagarakis2013compliant} uses compliant joints in the hip, knee, and ankle to store and release energy during locomotion.

\mypara{Control Strategies} Energy-efficient control strategies have also been explored. By removing the velocity tracking reward during post-disturbance training, a compliant reinforcement learning policy \cite{hartmann2024deep} leads to more energy-efficient push recovery. Similarly, COMAN \cite{kormushev2019learning} uses RL to optimize its center-of-mass trajectory for better energy efficiency.

Our work develops both hardware and software strategies to achieve energy-efficient locomotion. Our design features a near-symmetrical hip arrangement (front and back) with straight knees, which satisfies the static standing requirement for passive walking. Furthermore, we explore directly optimizing passivity within an end-to-end deep reinforcement learning framework to improve energy efficiency.

\section{Method}

\subsection{Hardware Design}

\begin{figure}[]
\includegraphics[width=0.975\linewidth]{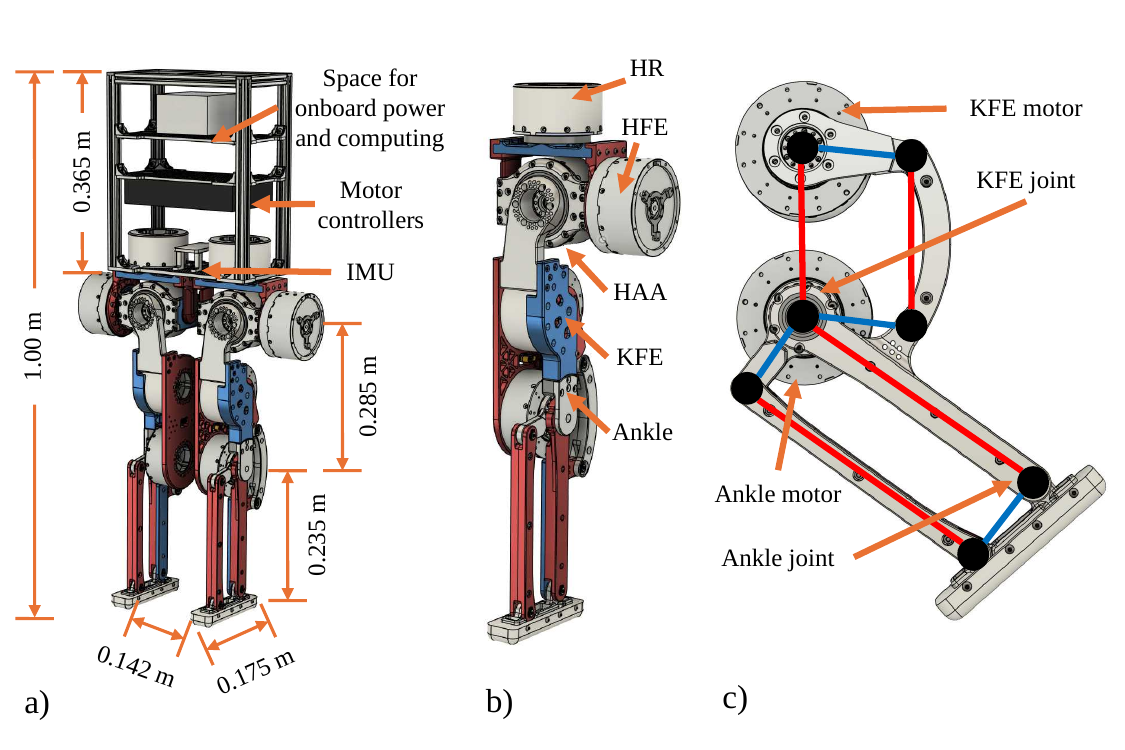}
\vspace{-5pt}
\caption{\textbf{Mechanical Design Overview:} a) Major dimensions and the extensible body design. b) All joints in the left leg. c) Two parallel linkages in the knee and ankle.}
\label{fig: body}
\vspace*{-6mm}
\end{figure}

The Duke Humanoid (Fig.~\ref{fig: body}) is a child-sized, open-source bipedal robot designed for dynamic locomotion. Standing $\SI{1}{\meter}$ from shoulder to foot and weighing $\SI{30}{\kilogram}$, it approximately matches the size of an average 8-year-old child \cite{growthcharts}. This size was selected for two reasons: first, it allows the robot to interact comfortably with human-centric environment surfaces while standing, enabling future object manipulation tasks; second, its mid-size balances functionality, affordability, and customizability, especially given the high cost of commercially available geared motors.

Following a co-design philosophy for complex robotic platforms \cite{li_design_2001}, we iteratively tuned and selected the mechanical design and actuators based on simulations.  We designed preliminary humanoid models in Isaac Gym and trained walking policies via reinforcement learning, using the resulting joint torque and velocity data to guide actuator selection.

\mypara{Mechanical Design} Our robot features 10 degrees of freedom (DoFs), with 5 DoFs in each leg: 3 in the hip, 1 for knee flexion and extension (KFE), and 1 for ankle plantar flexion and dorsiflexion. The DoFs for the hip include hip rotation (HR), hip flexion and extension (HFE), and hip abduction and adduction (HAA). The center of mass of the hip is aligned in the frontal plane, allowing the robot to maintain balance and walk with straight knees.

The proportions of our leg design anatomically match that of a real human. The tibia measures \SI{0.235}{\meter}, and the femur measures \SI{0.285}{\meter}, resulting in a tibia-to-femur ratio of 0.825, which is comparable to the average human ratio of 0.78 \cite{aitken2021legratio}. In addition, the range of motion of our leg joints largely covers that of a real human, as shown in Table~\ref{table:rangeofmotioncompare}.

The KFE joint is actuated by a motor via a parallel linkage. Similarly, the ankle motor, positioned at the KFE joint, connects to the ankle through another parallel linkage. As a result, the ankle joint is actuated in conjunction with the KFE and ankle motors, with a linear relationship between their movements. The parallel linkages relocate the motors higher in the leg, raising the leg's center of mass and reducing the lower leg's inertia, thereby reducing the torque required to actuate the KFE and hip motors. To save weight and simplify the design, the ankle does not include the inversion and eversion DoF. To compensate this DoF, the foot is designed with a linear contact surface to ensure consistent contact regardless of the roll angle.

\mypara{Actuators} Each joint is powered by a brushless DC motor with a planetary gearbox from Motorevo. The joint position is measured by a single-turn encoder at the back of the motor shaft. To simplify the design, the same brushless DC motor is used throughout, and varying reduction ratios are employed to satisfy the torque and velocity demands of each joint. Specifically, The HR, HAA, and KFE joints use 18:1 motors, providing $\SI[inter-unit-product=\cdot\mkern-1mu]{72}{\newton\meter}$ rated torque and $\SI{20}{\radian/\second}$ maximum velocity; the HFE joints use 20:1 motors, providing $\SI[inter-unit-product=\cdot\mkern-1mu]{80}{\newton\meter}$ rated torque and $\SI{18}{\radian/\second}$ max velocity; the ankles are powered by 10:1 motors, providing $\SI[inter-unit-product=\cdot\mkern-1mu]{40}{\newton\meter}$ rated torques and $\SI{36}{\radian/\second}$ max velocity. This uniform motor selection provides flexibility, as the motors share identical rotors and mounting configurations, making them interchangeable. Motion control is achieved through EtherCAT communication at \SI{2000}{\hertz} to minimize latency.

\begin{table}[t!]
\centering
\begin{tabular}{@{}llllll@{}}
\toprule
Joint  & HR & HAA & HFE & KFE & Ankle\\ \midrule
Human [deg] & [-50,40] & [-40,20] & [-30,110]& [0,150] & [-20,50] \\
Ours [deg] & [-90,60]& [-40,40]& [-90,90]& [0,110]& [-45,45]\\
Coverage & 100\%& 100\%& 85\%& 73\%  & 95\%\\ \bottomrule
\end{tabular}
\caption{\textbf{Range of Motion:} Our design mostly covers the range of motion of human leg joints \cite{liao2024berkeley}.}
\label{table:rangeofmotioncompare}
\vspace{-15pt}
\end{table}

\mypara{Manufacturing} The outer frame of the upper body is built using off-the-shelf aluminum extrusions, while the body plates and legs are custom-machined from aluminum 6061. This modular design allows for easy modifications. The upper body has enough space ($250 \times 120 \times \SI{200}{mm}$) reserved for future onboard power and computing to accommodate outdoor experiments. The body plates feature a grid of screw holes, allowing the attachment of custom 3D-printed electronic mounts. Additionally, the leg motors are covered with 3D-printed thermoplastic polyurethane (TPU) parts to protect against collisions.

\mypara{Sensors} An IMU (MicroStrain 3DM-CV7-AHRS),  mounted on the base plate,  provides angular velocity and orientation data. Two load cells are included at the toes, and two load cells are included at the heels. A Teesny 4.0 microcontroller interfaces with the IMU and the load cells and transmits the data to a computer through a serial connection.

\subsection{Reinforcement Learning for Bipedal Locomotion}

\mypara{Task and Observation} To demonstrate the locomotion capability of our platform, we develop a baseline control policy using RL. The task is to train the robot to follow a desired linear velocity $\mathbf{v}^*$ and angular velocity $\mathbf{\omega}^*$. Our policy observation includes the base angular velocity $\mathbf{v}_b$, the base gravity vector $\mathbf{g}$ projected onto the robot's base frame $\mathbf{g}_b$, joint positions $\mathbf{q}$, joint velocities $\dot{\mathbf{q}}$, and target body velocity commands, consisting of the linear velocities in the xy-plane and angular velocity around the z-axis, $[\mathbf{v}_{b,x}^*, \mathbf{v}_{b,y}^*, \mathbf{\omega}_{b,z}^*]$.

\mypara{Observations} Additionally, the observation also includes a periodic phase signal $\mathbf{\phi}$ encoded as $[\sin \mathbf{\phi}, \cos \mathbf{\phi}]$, and a desired foot contact pattern $c^*$ derived from the periodic phase signal. We employ an asymmetric actor-critic architecture \cite{chen2020learning}, where the critic network receives additional privileged information such as linear velocity ($\mathbf{v}_b$) and actual foot contact indicator ($c$). The actor and critic networks consist of three fully connected layers with hidden sizes of $[512, 256, 128]$ with ELU activation functions \cite{clevert2015fast}. 

\mypara{Actions} Our RL policy outputs $\mathbf{a}_q$ which is converted to target joint positions ($\mathbf{q}^*=\mathbf{q}_0+k_q\mathbf{a}_q$) at \SI{50}{\hertz}. Where $\mathbf{q}_0$ are the default joint positions, $k_q$ is a scaling factor for the action. The positions are smoothed with a first-order exponential low-pass filter before being converted to torques via a position PD controller, which operates at \SI{200}{\hertz} in simulation. In the real robot, the position PD controller runs at \SI{2000}{\hertz} to ensure smooth control. The PD gains are empirically determined to be $\SI[inter-unit-product=\cdot\mkern-1mu]{60}{\newton\meter\per\radian}$ and $\SI[inter-unit-product=\cdot\mkern-1mu]{5}{\newton\meter\second\per\radian}$.

\begin{table}[t]
% \vspace*{-15pt}
\centering
\resizebox{0.48\textwidth}{!}{%
\begin{tabular}{rcr}
\toprule
 Reward Terms & Definition & Weight \\
\midrule
Linear velocity & $\phi\left(\mathbf{v}_b^*-\mathbf{v}_b,-4*[1,1,0.1] \right)$& $1 $ \\
Angular velocity  & $\phi\left(\mathbf{\omega}_b^*-\mathbf{\omega}_b,-8*[0.1,0.1,1] \right)$& $0.5 $ \\
Base height  & $\phi\left(h_b^*-h_b,-2000 \right)$& $0.1$\\
 Orientation & $max \left( \left\|\mathbf{g}_{b,xy}\right\|^2, 0.1\right)$&$-20$\\
\midrule
% Joint power  & $\|\dot{\mathbf{q}}\cdot\mathbf{ \tau}\|_1$& $-0.0005$\\
Joint acceleration  & $\phi \left( \ddot{\mathbf{q}}, -10^{-4} \right)$ & $0.1 $ \\
Joint torque  & $\exp\left( G^{-1}\|\tau\|_1 \right)$& $0.05$\\
Joint limit & $ \left\|  \mathbf{q}  - \text{CLIP}\left( \mathbf{q},\mathbf{q}_{\min},\mathbf{q}_{\max}  \right)  \right\|^2$ & $-100$ \\
% Joint power & $\sum max \left( \tau \dot{\mathbf{q}},0 \right) $ & $0$ \\
% Joint power & $\| max \left( \tau \dot{\mathbf{q}},0 \right)\|_1 $ & $0$ \\

% Action rate  & $\left\|\mathbf{\dot{a}}\right\|^2$ & $-0.001$\\
% Feet stance time & $\sum_{k=0}^{n_f} (t_{stance,k} - 0.5)$ & $0.1 $ \\
\midrule
Feet air time & $\sum_{i=0}^{n_f} (t_{air,i} - 0.3)$& $1 $\\
Contact pattern& $1- {n_f}^{-1} \|c^*-c\|_1$&$0.5$\\
Feet orientation& $\phi \left(  \mathbf{g}_{f,xy},-8 \right)$&$0.1$\\
Feet forward& $\phi \left(  \mathbf{f}_{f,xy},-8 \right)$&$0.1$\\
Feet position& $\phi\left(\mathbf{p}_f^*-\mathbf{p}_f,[-1000,-1000,0] \right)$&$0.1$\\
Feet step height& $ \| max \left( h_f , h^*_f \right) \|_1$  &  $0.5$\\
\bottomrule
\end{tabular}
}
\caption{Baseline RL policy reward functions.}
\label{table:baseline_reward}
\vspace*{-15pt}
\end{table}

\mypara{Reward} The reward function (Table~\ref{table:baseline_reward}) includes terms for tracking and regularization (body, joint, and gait). Here, $\phi(\psi, w) := \exp(\sum_{i=1}^n w_i{\psi}_i^2)$ represents the exponential of the sum of the squared quantity $\psi$ weighted by $w$. $\mathbf{v}_b$, $\mathbf{\omega}_b$, $h_b$, and $\mathbf{g}_b$ represent the base linear velocity, angular velocity, height, and projected gravity. $\mathbf{q}_{\min}$ and $\mathbf{q}_{\max}$ represent the minimum and maximum joint position limits. $\mathbf{\tau}$ refers to joint actuation torque, $\ddot{\mathbf{q}}$ to joint accelerations, and $G$ to the robot's total gravity. 
% $\mathbf{\dot{a}}$ is the rate of action, 
$t_{\text{air}}$ is the duration the feet remain airborne, and ${n_f}=2$ is the number of feet. Finally, $\mathbf{g}_f$ and $\mathbf{f}_f$ are the feet's gravity and forward direction vectors, while $\mathbf{p}_f$ and $\mathbf{p}_f^*$ are the actual and desired feet positions, and $h_f$ and $h_f^*$ denote the actual and desired step heights.

\mypara{Training} We trained $4096$ parallel environments with Isaac Gym \cite{rudin2021learning} for \SI{30}{\minute} on a single NVIDIA RTX 4090 GPU.

\subsection{Policy for energy-efficient locomotion}
Energy efficiency in humanoid locomotion is important because the majority of the energy is used to actuate the joint during locomotion. Thus, in addition to our baseline RL policy, we integrate passive dynamics into the end-to-end reinforcement learning framework. We hypothesize that encouraging passive control in locomotion can enable a more energy-efficient controller. To achieve this, we propose explicitly modulating the joint torques to switch between passive and active modes. Specifically, we introduce a per-joint torque activation parameter, $\mathbf{\alpha}$, which scales existing position-based PD control. When $\mathbf{\alpha} = 0$, the joint torque will be scaled to zero, making the joint completely passive, allowing its movement to be governed by passive dynamics.  The modified control is expressed as:
\begin{equation*}
\tau =\mathbf{\alpha} \left( k_p ( \mathbf{q}^* - \mathbf{q}) + k_d (\dot{\mathbf{q}}^*- \dot{\mathbf{q}}) \right) 
\end{equation*}
\noindent where $k_p$ and $k_d$ are stiffness and damping factors.  $\mathbf{q}$ and $\mathbf{q}^*$ are the desired and actual joint position. $\dot{\mathbf{q}}^*=0$ is the desired joint velocity (always 0), and $\dot{\mathbf{q}}$ is the actual joint velocity.  The output of the new actor network is $[\mathbf{a}_q,\mathbf{ a}_\alpha] \in \mathbb{R}^{2 \times \text{DoF}}$. A sigmoid function constrains $\mathbf{ a}_\alpha \in [0,1] $ , where the constant $k$ is empirically set to 10. $\mathbf{ \alpha}_0 \in [0,1] $  is a constant that controls the minimum activeness of the scaling factor:
\begin{equation*}
\mathbf{\alpha}  =\mathbf{\alpha}_0 + (1-\mathbf{\alpha}_0)\cdot \text{sigmoid} \left( k \mathbf{ a}_\alpha \right)=\mathbf{\alpha}_0 +  \frac{(1-\mathbf{\alpha}_0)}{1+ e^ {-k \mathbf{a}_\alpha}} 
\end{equation*}

To encourage the use of passive actions, we introduce a passive action reward, defined as $r_{ \mathbf{a}_\alpha}=\|1-\mathbf{a}_\alpha\|_1$, scaled by $0.005$. In addition, to prevent the policy from getting stuck at small $\alpha$ values, we implement a linear curriculum for $\alpha_0$ to decay initially from $\alpha_0=0.5$ to $\alpha_0=0$.

\subsection{Bridging the Sim-to-Real Gap}

Transferring trained RL policies from simulation to the real robot, particularly for complex platforms such as humanoids, remains challenging. Several factors contribute to the sim-to-real gap: Motor backlash can cause discontinuities in actuation, sensor noise introduces uncertainty in observations, and dynamics mismatches, such as differences in joint friction, link mass, and inertia, can change the underlying transition probabilities.

To bridge these gaps, we first identified critical joint dynamics parameters that are tunable in simulation, including damping, friction, and armature. For each joint, we conducted control frequency sweeps between 1.5-3 \si{\hertz} to reduce discrepancies between simulation and real-world performance, specifically focusing on joint position, velocity, target position, and torque. This identification process enabled a closer match between the robot’s simulated and real-world behavior.

We also applied domain randomization to increase the robustness of the simulation policy for real-world transfer. Table \ref{table:randomization} outlines the key parameters. Mass and inertia randomizations compensate for inconsistencies introduced by moving cables, electronics, loads, and mechanical model inaccuracies. Joint offset randomizations address variations in default joint positions after power-up and calibration. Position, velocity, and motor strength randomizations mitigate the effects of motor backlash, encoder noise, and actuation inconsistencies. Additionally, angular velocity and gravity vector randomizations simulate IMU noise, and friction randomizations consider locomotion on different surfaces.

\begin{table}
\centering
\begin{tabular}{llll}
\toprule
Parameter & Unit   & Range & Operator \\
\midrule
Base COM            & \si{\meter}              & $[-0.02,0.02]$     & additive         \\
Base mass           & \si{\kilogram}             & $[-0.5, 5]$        & additive         \\
Link mass           & \si{\kilogram}             & $[0.95, 1.05]$    & scaling          \\
Link inertia        & \si[inter-unit-product = \ensuremath{\cdot}]{\kilogram\meter\squared}  & $[0.95, 1.05]$    & scaling          \\
Joint offset        & \si{\radian}           & $[-0.02, 0.02]$   & additive          \\
Joint position      & \si{\radian}           & $[-0.03, 0.03]$   & additive          \\
Joint velocity      & \si{\radian/\second}         & $[-1.5, 1.5]$     & additive          \\
Motor strength      & -             &  $[0.98, 1.02]$       & scaling       \\
Observation delay   & \si{\milli\second}            & $[0, 20]$         & -                 \\
Angular velocity    & \si{\radian/\second}         & $[-0.2, 0.2]$     & additive          \\
Gravity vector      & -             & $[-0.05, 0.05]$   & additive          \\
Friction            & -             & $[0.2, 1.2]$      & -                 \\
% PD Factors          & \%            & [80, 120]       & scaling           \\ \hline
\bottomrule
\end{tabular}
\caption{Domain randomization parameters.}
\label{table:randomization}
\vspace{-5pt}
\end{table}

\section{Experiments and Results}

This section demonstrates the performance of our baseline locomotion policy in a zero-shot sim-to-real setup on our humanoid platform. Additionally, we evaluate the real-world performance of our passive walking policy. To further validate the effectiveness of our passive walking policy, we compare its energy efficiency and gait characteristics against the baseline policy in simulation and reality.

\begin{figure}[t!]
\begin{center} 
\includegraphics[trim={ 0  0  0 2cm},clip,width=0.8\linewidth]{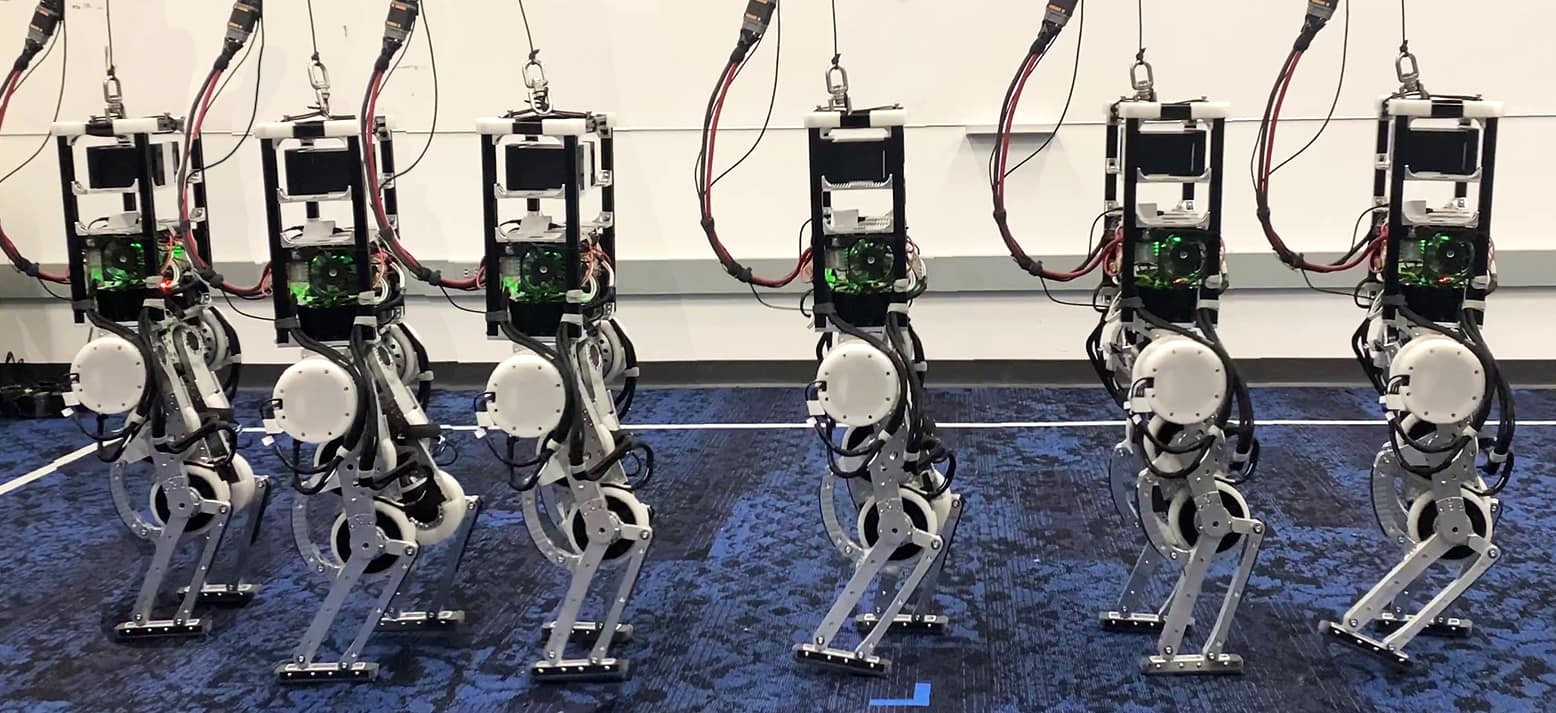}
\end{center}
\vspace{-5pt}
\caption{Chronophotograph showing the Duke Humanoid walking using the baseline RL policy.}
\label{fig: sim2real}
\vspace{-15pt}
\end{figure}

\subsection{Locomotion in the Real}
%To evaluate the transferability of our simulation policy to the real robot, 
We deployed our baseline policy zero-shot to our humanoid hardware. Fig.~\ref{fig: sim2real} shows a chronophotograph of the robot walking forward on a flat carpeted surface at a command velocity of \SI{0.3}{\meter/\second}.

\begin{figure}[t!]
    \centering
    \vspace{4pt}
    \includegraphics[width=1\linewidth]{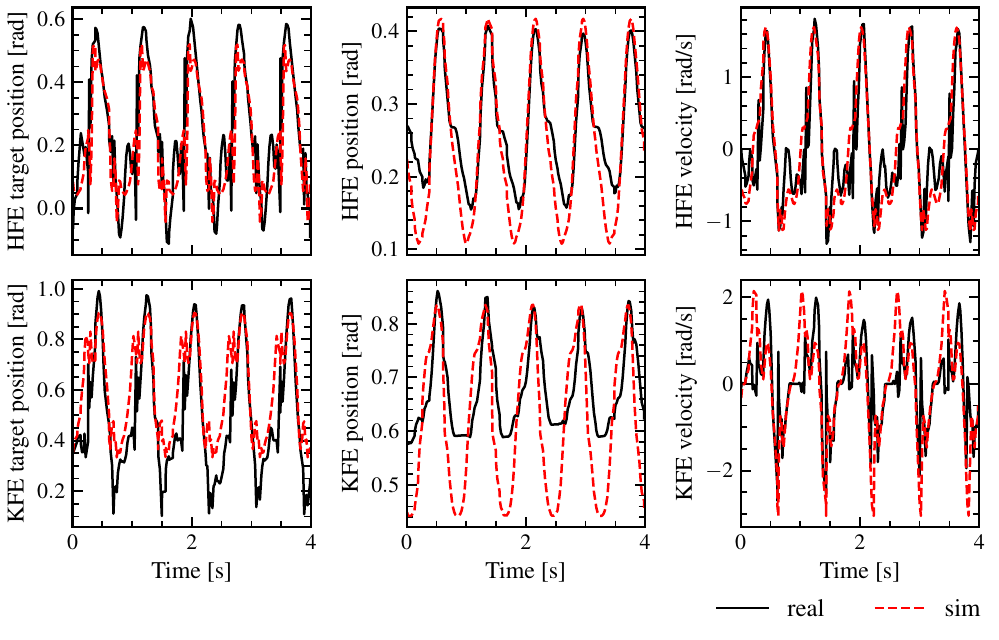}
    \vspace{-20pt}
    \caption{\textbf{Comparison of Simulated and Real-World Baseline Walking}: Target joint positions (from RL policy), actual joint positions, and joint velocities for the HFE and KFE joints.}
    \vspace{-5pt}
    \label{fig:baselin_sim2real}
\end{figure}

As shown in Fig.~\ref{fig:baselin_sim2real}, we quantitatively compared the target joint positions, the actual joint positions, and the joint velocities of the HFE and KFE joints between simulation and real-world performance. We chose to focus on these joints because they are critical for forward movements. We aligned the first steps of the simulation with the real data and compared the relative target joint positions. The maximum difference in the target joint position is \SI{0.1}{\radian} for HFE and \SI{0.15}{\radian} for KFE, indicating that the policy produces similar actions in the simulation and the real world. For the actual joint positions, the maximum error is \SI{0.1}{\radian} for HFE and \SI{0.2}{\radian} for KFE, which is only significant during half of the gait cycle. Overall, the differences between the target joint positions and joint velocities are bounded, resulting in a similar behavior between the simulation and the real system.

\vspace{-10pt}

\begin{figure}[t!]
    \centering
    \includegraphics[width=1\linewidth]{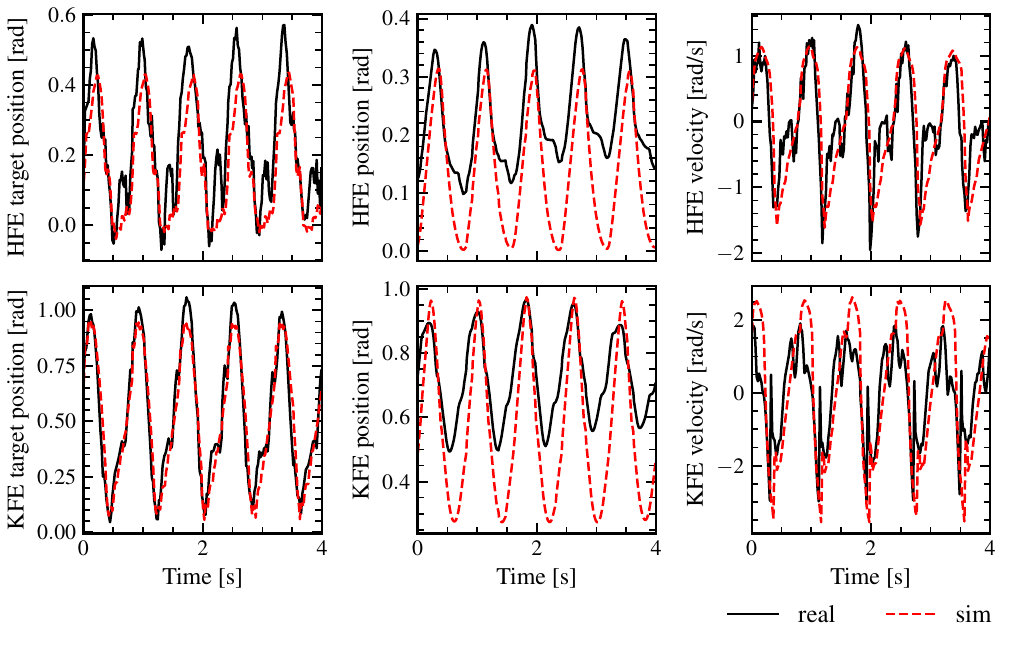}
    \vspace{-20pt}
    \caption{\textbf{Comparison of Simulated and Real-World Passive Walking}: Target joint positions (from RL policy), actual joint positions, and joint velocities for the HFE and KFE joints.}
    \vspace{-15pt}
    \label{fig:passive_sim2real}
\end{figure}

\vspace{10pt}
\subsection{Passive Policy for Energy-Efficient Locomotion}

The trained passive policy behaves visually differently compared to the baseline policy at zero velocity. The baseline policy actively steps in place, whereas the passive policy remains motionless, enhancing energy efficiency.

\mypara{Energy Efficiency} A key metric for quantifying the energy efficiency of robot locomotion is the Cost of Transport (CoT). A lower CoT translates to higher energy efficiency. CoT is defined as the energy required for locomotion, normalized by the distance traveled and the robot weight \cite{zarrouk2013cost}:  
\begin{equation*}
% L(d,\tau_i,\omega_i) = \frac{\sum_{i=0}^{9}\tau_i\omega_i}{mgd} 
CoT = \frac{W_{input}}{m\cdot g\cdot d}
\label{equation:cot}
\end{equation*}

\noindent where $m\cdot g$ is the robot's total gravity, $W_{input}$ is the energy input, and $d$ is the measured distance traveled. Ideally, the $W_{input}$ is obtained by measuring the electrical power usage at the power supply, however, this term cannot be directly measured in simulation. In the context of legged robots, the majority of the power comes from joint actuation. Therefore, we approximate the energy input by integrating the motor's mechanical power.

\begin{figure}[!t]
    \centering
    \vspace{4pt}
    \includegraphics[width=1\linewidth]{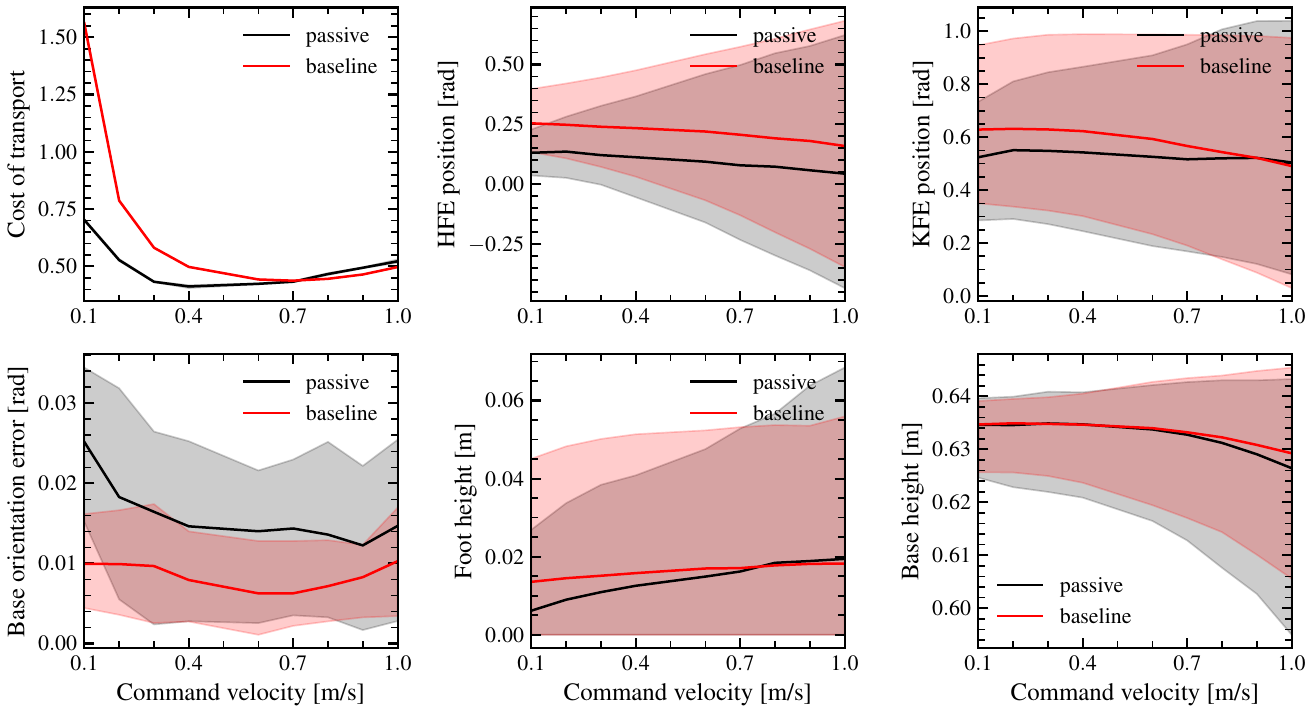}
    \vspace{-15pt}
    \caption{\textbf{Comparison of baseline vs. passive policy for walking in simulation}: Red/grey shaded regions represent the 95\% confidence interval. The passive policy exhibits up to 50\% better energy efficiency at low speed, with reduced KFE and HFE positions, indicating less action in walking.}
    \label{fig:cot_walk}
    \vspace{-15pt}
\end{figure}

\begin{figure}[!t]
    \centering
    % \vspace{4pt}
    \includegraphics[width=1\linewidth]{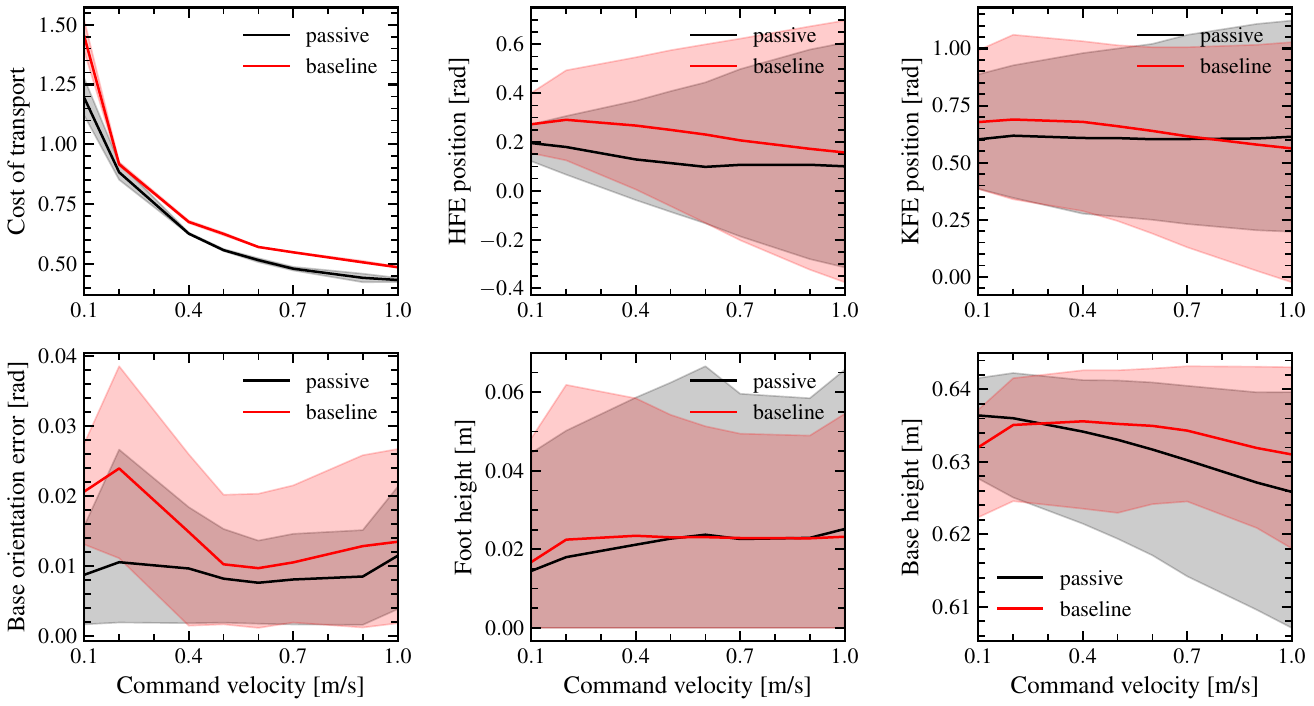}
    \vspace{-15pt}
    \caption{\textbf{Comparison of baseline vs. passive policy for running in simulation}: Red/grey shaded regions represent the 95\% confidence interval. The passive policy generally outperforms the baseline by 10\% and exhibits a lower HFE position. }
    \label{fig:cot_run}
    \vspace{-15pt}
\end{figure}

To evaluate the effectiveness of our passive control algorithm, we trained a passive policy for walking (stance ratio of $0.6$) and running (stance ratio of $0.38$) in the simulation. We then compared the passive policy with its active counterpart across a range of command velocities, from \SI{0.1}{\meter/\second} to \SI{1}{\meter/\second}. As shown in Fig.~\ref{fig:cot_walk}, at low walking speeds, from \SI{0.1}{\meter/\second} to \SI{0.5}{\meter/\second}, the passive policy outperforms the baseline in CoT, achieving up to a $50\%$ reduction at \SI{0.1}{\meter/\second}. The range of motion for both HFE and KFE is lower under the passive policy at low velocities. Base height (the IMU's distance from the ground) and foot heights are similar between the baseline and passive policies, indicating comparable gait patterns. The passive policy exhibits a larger base orientation error for walking, averaging \SI{0.01}{\radian}, suggesting that the robot maintains a forward tilt to take advantage of passive dynamics. For running, as shown in Fig.~\ref{fig:cot_run}, the passive policy also outperforms the active baseline in running, albeit by a smaller margin of $10\%$. The range of motion for HFE is reduced compared to the baseline policy; the base height variability is increased, suggesting a shift in the center of mass. Other metrics remain comparable between the passive and baseline policy.

\begin{figure}[h!]
    \centering
    \includegraphics[width=0.9\columnwidth]{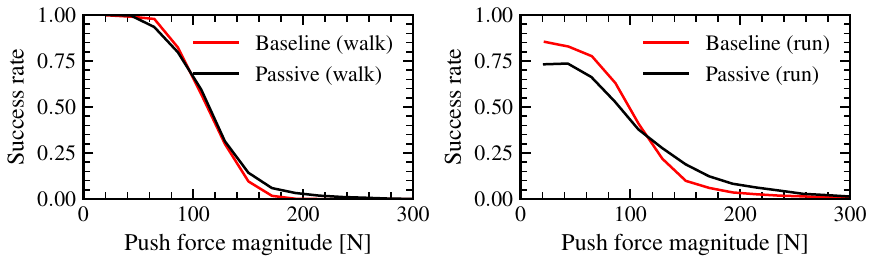}
    \vspace{-5pt}
    \caption{\textbf{Push recovery comparison}: baseline vs. passive policy for walking (left) and running (right) in simulation.}
    \vspace{-15pt}
    \label{fig:push_rejection}
\end{figure}

\mypara{Robustness} We evaluated the robustness of the learned policy by measuring the success rate of push recovery in simulation. In each trial, a random horizontal force sampled uniformly within a disk of \SI{300}{\newton} is applied to the humanoid base body between \SI{2}{}-\SI{3}{\second} for a duration of \SI{0.1}{\second}. Push recovery is considered successful if the robot does not fall or self-collide for \SI{5}{\second}. Fig.~\ref{fig:push_rejection} compares the push recovery success rates of the baseline and passive policies trained for walking and running. The passive walking policy achieves similar success rates in push recovery as the baseline walking policy, while the passive running policy performs at a slightly reduced success rate at lower push magnitude, and slightly better in higher magnitude push recovery. The result indicates that the robustness of the control policy remains mostly unchanged with the addition of passive actions.

% \FloatBarrier

% \begin{figure}[h!]
% \centering
% \includegraphics[width=0.8\linewidth]{figures/chrono_passive_v1.2.jpg}
% \vspace{-4pt}
% \caption{Chronophotograph showing the humanoid walking using the passive policy at a command velocity of 0.3 m/s.}
% \label{fig: passive2real}
% \vspace{-15pt}
% \end{figure}
% \vspace{5pt}

\begin{figure}[h!]
    \centering
    \vspace{5pt}
    \includegraphics[width=1\linewidth]{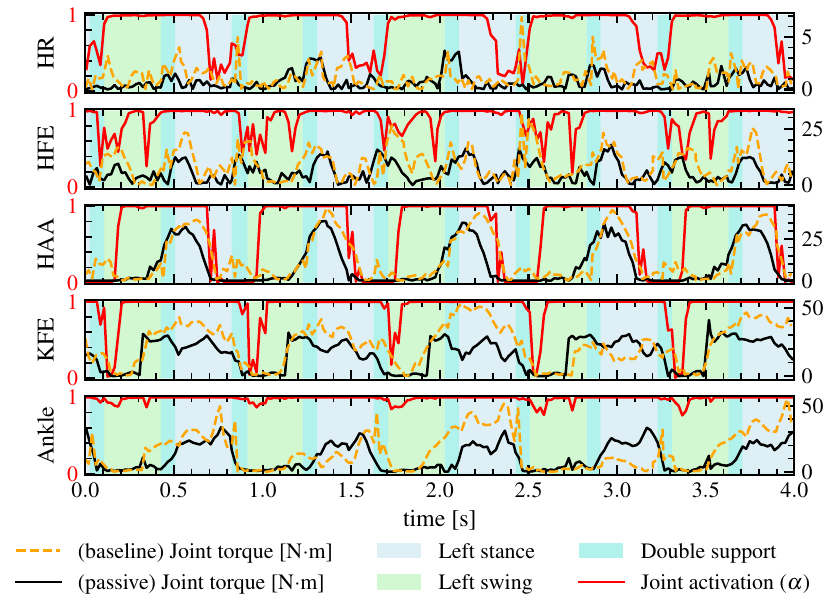}
    \vspace{-18pt}
    \caption{\textbf{Comparison of left leg absolute joint torque for real-world baseline vs. passive walking}: KFE deactivation during swing in passive walking demonstrates the utilization of passive dynamics, contrasting with the generally higher torque of the baseline policy.}
    \label{fig:passive_activation}
    \vspace{-15pt}
\end{figure}

\subsection{Passive Locomotion in the Real}

We deployed the passive control policy zero-shot on our humanoid platform. % Fig.~\ref{fig: passive2real} shows it walking forward at a command velocity of \SI{0.3}{\meter/\second}. %
The supplementary video provides a visual comparison of the walking performance baseline and passive policies. As shown in Fig.~\ref{fig:passive_sim2real}, the target and actual joint positions and velocities for the HFE and KFE joints suggest a reasonable agreement between simulation and real-world execution. This observation aligns with the sim-to-real performance observed for the active baseline (Fig.~\ref{fig:baselin_sim2real}).

To visualize the joint actuation pattern, we show the single (left) leg dynamics of controlled joint activation and absolute torque during passive walking (Fig.~\ref{fig:passive_activation}). At a command velocity of \SI{0.3}{\meter/\second}, all hip joints exhibit periodic activation patterns. The KFE deactivates during leg swing, suggesting the use of passive dynamics for leg swing. The ankle joint does not display a significant periodic activation pattern, likely because high joint friction reduces the lower leg's tendency to passively rotate. We confirmed similar periodic patterns for the right leg. Note that the joint actuation pattern does not exactly match the human pattern because the humanoid does not exactly match human morphology such as mass distribution. 

We compared the real-world CoT of the passive policy with the baseline policy in three repeated trials at a command forward velocity of \SI{0.3}{\meter/\second}. The baseline policy achieved an average CoT of $1.13 \pm 0.1$, and the passive policy improved the CoT to $0.77\pm0.1$, resulting in a $31\%$ CoT reduction. Both simulation (Fig.~\ref{fig:cot_walk}) and real-world results show that the passive policy outperforms the baseline in energy efficiency during low-speed walking.

\section{Conclusion, Limitation, and Future Work}

We introduce the Duke Humanoid v1.0, an open-source hardware and software research platform for studying humanoid locomotion. This paper details our design choices and the development of a baseline reinforcement learning policy for bipedal locomotion. Furthermore, we explore directly modulating joint activation within the end-to-end reinforcement learning framework to leverage passive dynamics for energy-efficient humanoid walking. Preliminary real-world experiments demonstrate that our approach achieved a $31\%$ reduction in the cost of transport.

However, our platform currently lacks arms, limiting our ability to study whole-body control and mobile manipulation. In the future, we plan to equip the robot with arms and dexterous hands. Our robot is currently tethered, limiting its mobility.  We plan on implementing onboard power to study more dynamic motions, such as running and jumping.

\FloatBarrier  % Optional: Prevent figures from moving past this point

\bibliographystyle{IEEEtran}
\bibliography{main_paper}

\end{document}